\documentclass[10pt,twocolumn,letterpaper]{article}

\usepackage{iccv}
\usepackage{times}
\usepackage{epsfig}
\usepackage{graphicx}
\usepackage{amsmath}
\usepackage{amssymb}

\usepackage{multirow}
\usepackage{comment}
\usepackage[pagebackref=true,breaklinks=true,letterpaper=true,colorlinks,bookmarks=false]{hyperref}

\iccvfinalcopy % *** Uncomment this line for the final submission

\def\iccvPaperID{9319} % *** Enter the ICCV Paper ID here

\ificcvfinal\fi

\begin{document}

%%%%%%%%% TITLE
\title{ORC: Network Group-based Knowledge Distillation using Online Role Change}

\author{Junyong Choi$^{1,2}$, Hyeon Cho$^{1}$, Seokhwa Cheung$^1$, and Wonjun Hwang$^{1,3}$\\
$^1$Ajou University, Korea, $^2$Hyundai Motor Company, $^3$Naver AI Lab\\
{\tt\small chldusxkr@hyundai.com, \{ch0104, shjeong008, wjhwang\}@ajou.ac.kr}
}

\maketitle
\ificcvfinal\thispagestyle{empty}\fi

%%%%%%%%% ABSTRACT
\begin{abstract}
In knowledge distillation, since a single, omnipotent teacher network cannot solve all problems, multiple teacher-based knowledge distillations have been studied recently. However, sometimes their improvements are not as good as expected because some immature teachers may transfer the false knowledge to the student. In this paper, to overcome this limitation and take the efficacy of the multiple networks, we divide the multiple networks into teacher and student groups, respectively. That is, the student group is a set of immature networks that require learning the teacher's knowledge, while the teacher group consists of the selected networks that are capable of teaching successfully. We propose our online role change strategy where the top-ranked networks in the student group are able to promote to the teacher group at every iteration. After training the teacher group using the error samples of the student group to refine the teacher group's knowledge, we transfer the collaborative knowledge from the teacher group to the student group successfully. We verify the superiority of the proposed method on CIFAR-10, CIFAR-100, and ImageNet which achieves high performance. We further show the generality of our method with various backbone architectures such as ResNet, WRN, VGG, Mobilenet, and Shufflenet.\footnote{Our code is available at \href{https://github.com/choijunyong/ORCKD}{https://github.com/choijunyong/ORCKD}}
\end{abstract}

%%%%%%%%%%%%%%%%%%%%%%%%%%%%%%%%%%%%%%%%%%%%%%%%%%%%%%%%%%%%%%%
\begin{figure*}
\centering
\includegraphics[width=0.7\linewidth]{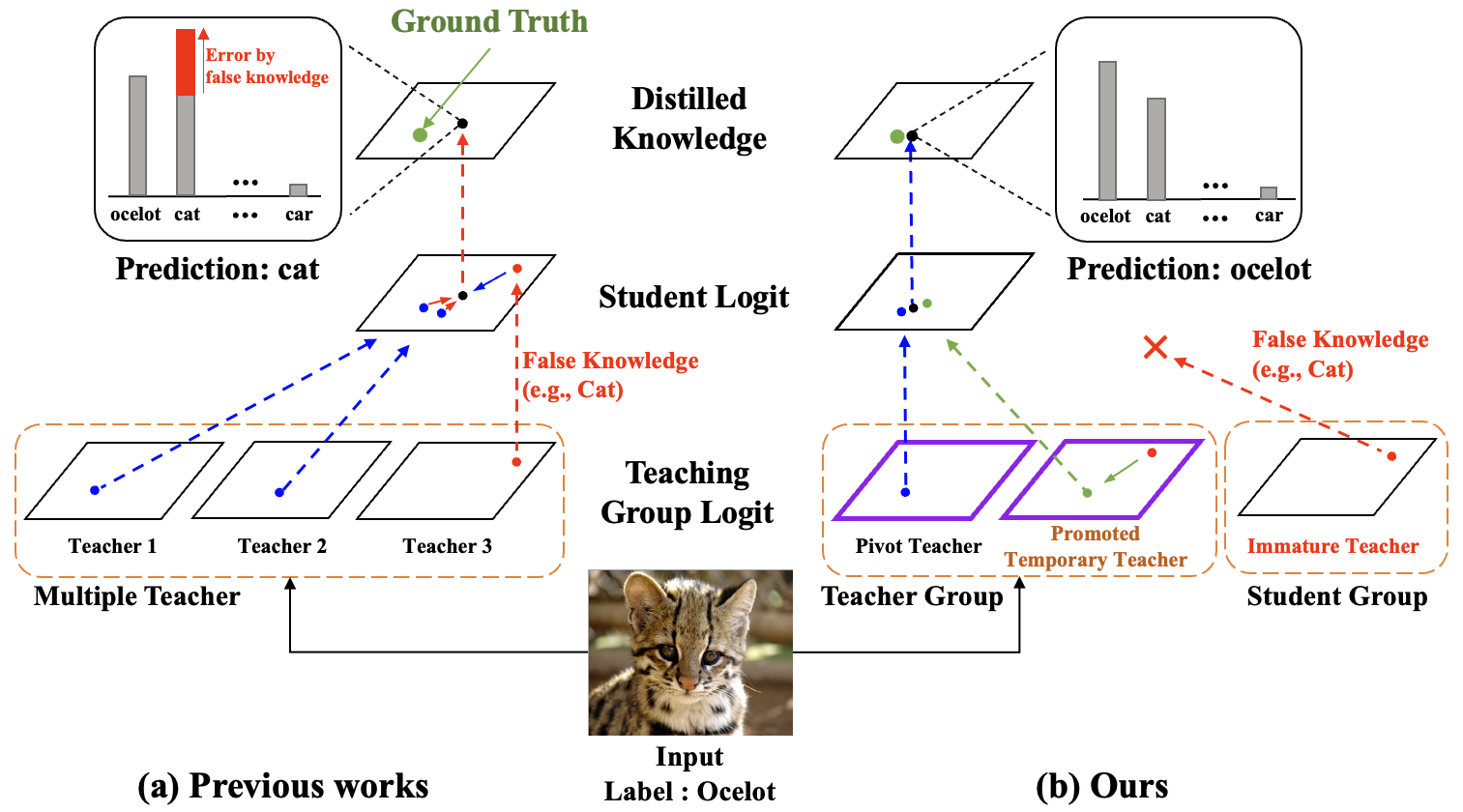}

\caption{\textbf{Network Group-based Knowledge Distillation.} (a) Multiple networks are still suffering from the problem of transferring false knowledge from the immature network. (b) Our method divides multiple networks into a teacher group and a student group according to the performance and assigns different roles to them every iteration. In each iteration, the role of the network belonging to each group can be continuously changed according to its changed performance. }
\label{fig:fig1}
\vspace{-2mm}
\end{figure*} 
%%%%%%%%%%%%%%%%%%%%%%%%%%%%%%%%%%%%%%%%%%%%%%%%%%%%%%%%%%%%%%%

%%%%%%%%% BODY TEXT
\section{Introduction}
\label{sec:intro}

Deep learning using convolutional neural networks (CNN) is making significant progress in computer vision tasks (e.g., object detection, classification, segmentation). For making the efficient network, many trials have been studied from quantization~\cite{wu2016quantized,chen2016quantized,rastegari2016quantizedr,hubara2016quantized} to pruning~\cite{he2018pruning,li2016pruning,han2015deeppruning,han2015learningpruning}. 
After Hinton's proposal~\cite{HintonKD} on Knowledge Distillation (KD), KD methods have been proposed in various forms~\cite{FitNets,FSP,CRD,Tung2019,SRRL, itkd, cho2021deep} but most of them leverage a pair of a single large teacher and a small student for knowledge transfer. However, there is a clear limitation to improving the performance of the student successfully because the teacher network is egocentric and complacent in spite of its good performance and note that the teacher is independently trained in advance without the consideration of the student's characteristics. 
Eventually, the teacher's knowledge is transferred from the viewpoint the teacher has learned in advance, not the direction in which the student can learn successfully, and the student network easily overfits in an undesired direction. Another limitation of KD is that when the difference of network sizes between the teacher and student networks is large, a single teacher-based KD does not properly transfer the teacher's knowledge to the student as described in~\cite{TAKD2020,DGKD}. 

Recently, the multiple teacher-based KD methods~\cite{TAKD2020,DGKD,Mutual,Quishan2020} have been proposed to solve the issues mentioned above. Specifically, Teacher Assistant-based KD (TAKD)~\cite{TAKD2020,DGKD} was proposed as a method of teaching students using multiple networks with different network sizes and online distillation methods~\cite{Mutual,one,Quishan2020} simultaneously learned the knowledge from the multiple networks of the similar capacity from the beginning. 
Most of all, the fundamental problem of the multiple network-based KD is that, as shown in Fig.~\ref{fig:fig1} (a), the networks are trained by using even knowledge from immature networks. As a result, the collaborative knowledge that should be used for learning can be contaminated by the false knowledge. On the other hand, we use multiple networks of different sizes to overcome the teacher-student large gap in online KD and divide the networks into a teacher group and a student group to prevent false knowledge from being used for learning as shown in Fig.~\ref{fig:fig1} (b).

In this paper, we propose a method to prevent the transfer of false knowledge. False knowledge refers to flawed information present in teacher networks, and our objective is to prevent the student networks from learning in the wrong direction by avoiding the assimilation of this false knowledge. We first deploy the multiple networks into the teacher group and the student group according to their roles. The main role of the teacher group is to educate the student network using collaborative knowledge, and the main role of the student group is to receive useful information from the teacher group and train optimally. What we need to know at this point is that group members' role could be changed according to their current performances on-the-fly during training and we call it \textbf{\textit{Online Role Change (ORC) strategy}}.  
In detail, the highest-performing student in the student group is promoted to the teacher group based on its demonstrated teaching ability at every iteration.
To prevent transferring false knowledge, ORC consists of three steps: Intensive teaching, Private teaching, and Group teaching. 
In intensive teaching, we create feedback samples for students' incorrect predictions and train pivot teacher using them so that it can focus on intensively about their incorrect information.
In order to narrow the gap with pivot teacher by further improving the teaching ability of promoted temporary teacher, private teaching is conducted by pivot teacher.
Through the previous two steps, the teacher group grows into a more complementary group and group teaching is conducted to guide the student group using their collaborative knowledge.
Our major contributions are summarized as follows:\\
\begin{itemize}
    \item We propose the novel multiple network-based KD using ORC mechanism that effectively prevents false knowledge transfer by promoting the top-ranked student network to a temporary teacher model during training.
    \item We suggest three teaching methods such as intensive, private, and group teachings to achieve the successful ORC for online KD.
    \item We promisingly show that the proposed method outperforms various well-known KD methods in CIFAR-10, CIFAR-100, and ImageNet.
\end{itemize}

%%%%%%%%%%%%%%%%%%%%%%%%%%%%%%%%%%%%%%%%%%%%%%%%%%%%%%%%%%%%%%%
% Figure 2 Main Framework
\begin{figure*}[t]
\centering
\includegraphics[width=0.9\linewidth]{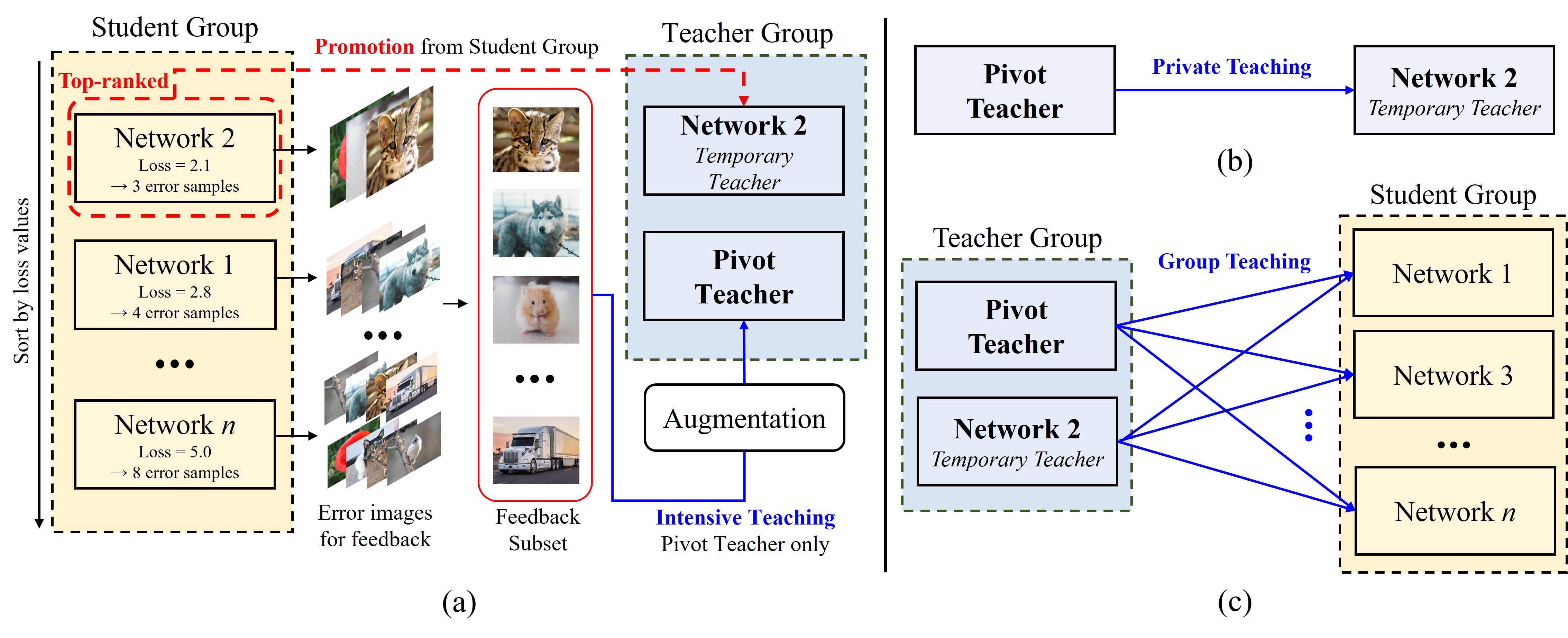}
%\vspace{-0.2cm}
\caption{\textbf{Overall framework of the proposed method.} \textbf{(a) Intensive teaching:} Mini-batch data is provided to all networks belonging to student group, and the number of feedback instances is determined based on the loss to construct a feedback subset. Simultaneously, according to the loss, the top-ranked network is promoted to the teacher group to perform the role of a temporary teacher. The feedback subset goes into the input of the pivot teacher after augmentation and proceeds with intensive teaching. \textbf{(b) Private teaching:} The pivot teacher teaches the best performing temporary teacher in the mini-batch privately. \textbf{(c) Group teaching:} A teacher group teaches a group of students. For reference, all of these processes are carried out in each iteration until the last epoch.}
\vspace{-2mm}
\label{fig:overall}
\end{figure*}
%%%%%%%%%%%%%%%%%%%%%%%%%%%%%%%%%%%%%%%%%%%%%%%%%%%%%%%%%%%%%%%

\section{Related Works}
We will describe two major categories of related works: KD based on a single teacher and multiple networks.

%\noindent
\textbf{KD based on a single teacher.} 
KD approaches for transferring knowledge from a pre-trained teacher network to a student network have been studied for many purposes such as reducing the computational complexity or transferring the core knowledge. The key to the basic KD approach was to mimic the knowledge (e.g., softened logits) extracted from the teacher network; Hinton \textit{et al.}~\cite{HintonKD} designed the first concept of knowledge distillation.
After that, Romeo \textit{et al.}~\cite{FitNets} proposed the method that allowed the student network to mimic the teacher network's intermediate hidden layers.
Zagoruyko \textit{et al.}~\cite{Zagoruyko2017} used attention maps to efficiently enable students to learn teacher's knowledge among layers.
Yim \textit{et al.}~\cite{FSP} defined the Flow of the Solution Procedure (FSP) matrix as knowledge by computing the inner product of two layers' feature maps.
Tung \textit{et al.}~\cite{Tung2019} introduced similarity-preserving knowledge distillation, in which input pairings that led to similar activation in the teacher also produced a similar activation in the student.
Park \textit{et al.}~\cite{RKD} advocated defining the relations (e.g., distance-wise, angle-wise) between the outputs of the teacher and student network as knowledge.
Tian \textit{et al.}~\cite{CRD} used contrastive-based objective for transferring knowledge between deep networks in order to maximize mutual information between two networks.
Xu \textit{et al.}~\cite{SELF2020} demonstrated that using contrastive learning as a self-supervision task allowed a student to gain a better understanding of a teacher network.
In this respect, many KD methods have been conducted on methods that allowed the student to learn successfully from one teacher, but there were still limitations in that the student network could not fully learn from the teacher due to the large gap between the teacher and the student networks or the overfitting that learned even the teacher's inherent errors.

%\noindent
\textbf{KD based on multiple networks.} 
When the student model's capacity was not enough to imitate the teacher model due to the large parameter gap, Cho \textit{et al.}~\cite{cho2019} observed that knowledge distillation could not work well and suggested a strategy to address this problem by terminating teacher training early to regain unripe knowledge more suitable to the student network. 

Another trial to alleviate it was TAKD proposed by Mirzadeh \textit{et al.}~\cite{TAKD2020} where they reduced the model capacity gap between teacher and student networks through sequentially connecting assistant networks and their model capacities were midway sizes among teacher and student networks.
Recently, Son \textit{et al.}~\cite{DGKD} observed an error-avalanche issue where the initial error of the teacher network increased widely through the sequential connected assistant networks, affecting the final student network. To solve this problem, they proposed the Densely Guided Knowledge Distillation (DGKD) where the whole multiple networks were densely connected to the student network.
On the other hand, to overcome the overfitting to the sole teacher network, many studies based on online learning have been proposed.
For example, Zhang \textit{et al.}~\cite{Mutual} presented Deep Mutual Learning (DML), in which students exchanged knowledge with each other during the training process without a well-trained teacher.
After this study, Lan \textit{et al.}~\cite{one} proposed On-the-fly Native Ensemble (ONE), which constructs gated ensemble logits of the training networks to enhance target network learning.
Guo \textit{et al.}~\cite{Quishan2020} suggested online Knowledge Distillation based on Collaborative Learning (KDCL) without a pre-trained teacher. They attempted to generate a soft target that improved all students, even if there was a capacity gap among students.
Chen \textit{et al.}~\cite{Defang2020} introduced a two-level distillation framework using multiple auxiliary peers and a group leader during training.

In the end, KDs are largely divided into two parts. One is concerned about how to increase learning efficiency from the viewpoint of the student network, and the other is concerned about how to effectively create the good knowledge to be taught from the viewpoint of the teacher network. 
In this paper, we basically belong to the latter and leverage the group's network-based knowledge distillation using online role change. We separate the teacher group and the student group, not using the whole networks like KDCL~\cite{Quishan2020}, and the top-ranked network of the student group can be promoted to the teacher group on-the-fly, which helps to avoid transferring false knowledge from the teacher.

\section{Online Role Change-based Group Network for Knowledge Distillation}
In this section, we provide a quick overview of KD~\cite{HintonKD} background. We detail our ORC and three teaching methods such as Intensive, Private, and Group teachings.
\subsection{Background}
The key concept of KD is extracting and transferring core knowledge from a larger teacher network to a smaller student network to mimic the softened class probability of a teacher network.
The framework of KD can be explained as follows:
Let $z_T$ and $z_S$ be logits of teacher and student networks, respectively, then each network’s final output of class probability would be $P_T$ and $P_S$ defined as follows:
\begin{equation}
    P_T= softmax(\frac{z_T}{\tau}),\quad P_S=softmax(\frac{z_S}{\tau}),
    \label{eq:01}
\end{equation}
where $\tau$ is the temperature parameter controlling the softening of the class probability. The Kullback-Leibler (KL) divergence is used to calculate the loss of KD. Original supervised loss ($L_{CE}$), the cross-entropy ($\mathcal{H}$), and knowledge distillation loss ($L_{KD}$) can be explained as follows:
\begin{equation}
\begin{aligned}
    L_{CE} = \mathcal{H}(&y,P(x)),\\ 
    \quad L_{KD} = \tau^2 KL(&P_S(x),P_T(x)),
\end{aligned}
\label{eq:02}
\end{equation}
where $y$ is a one-hot vector label, $P(x)$ is the class probability distribution: $P(x)=softmax(z)$. The student's total loss ($L_{S}$) is a combination of the supervision cross-entropy loss $L_{CE}$ and the KD loss $L_{KD}$, as described below:
\begin{equation}
    L_{S}=  (1-\beta)L_{CE} + \beta L_{KD},
    \label{eq:03}
\end{equation}
where $\beta$ is the balance parameter of $L_{CE}$ and $L_{KD}$.

\subsection{Proposed Method}

We propose the multiple network-based KD in an online manner like KDCL~\cite{Quishan2020} where the multiple network sizes are different to bridge the parameter gap between the teacher and student networks~\cite{TAKD2020}. When training in an online manner, it is possible to deliver knowledge that is easier for students to resemble by reducing the capacity gap between teachers and students. Also, DGKD~\cite{DGKD} has observed that the student network's performance did not improve significantly due to the transfer of false knowledge of some teachers in multiple network-based KD. As shown in the previous work of Fig.~\ref{fig:fig1} (a), it shows that the student is constantly confused in the process of training due to the teacher's false knowledge. The overall training framework can be confirmed through Fig.~\ref{fig:overall}.

\begin{figure}
\centering
\resizebox{\columnwidth}{!}{%
\begin{tabular}{c}
\includegraphics[width=\columnwidth]{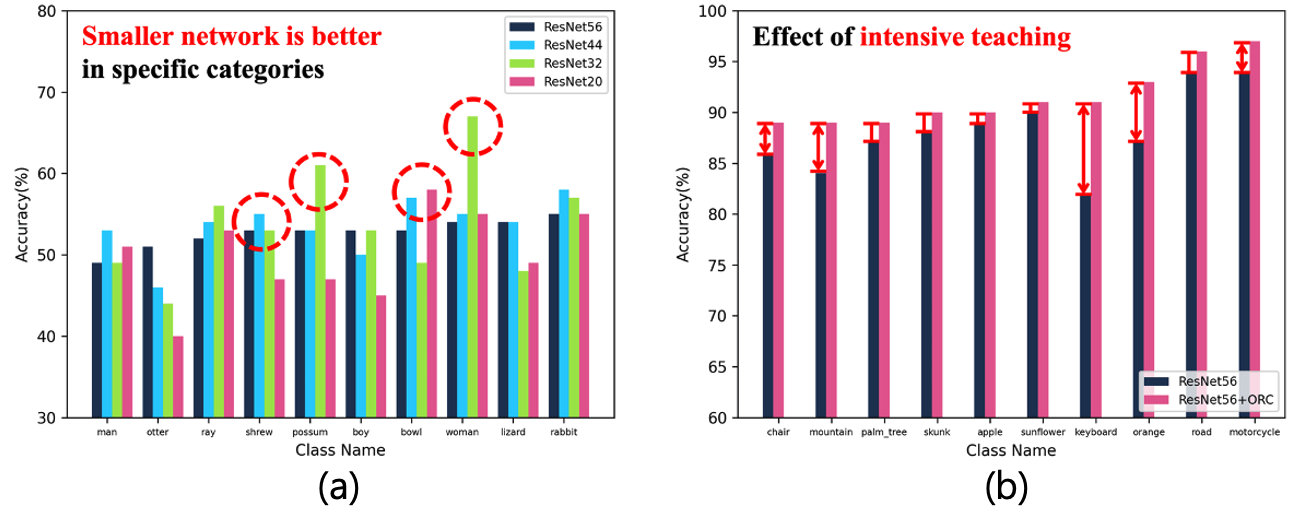}
    \end{tabular}
    }
% \vspace{-4mm}
   \caption{(a): Accuracy comparison of networks for difficult classes, (b): Accuracy improvement for difficult samples through intensive teaching.}
   % (c): Accuracy improvement of student groups on difficult samples after ORC.} 
   \label{fig:fig3}
% \vspace{-6mm}   
\end{figure}

\begin{figure}
\centering
\resizebox{\columnwidth}{!}{%
\begin{tabular}{c}
\includegraphics[width=\columnwidth]{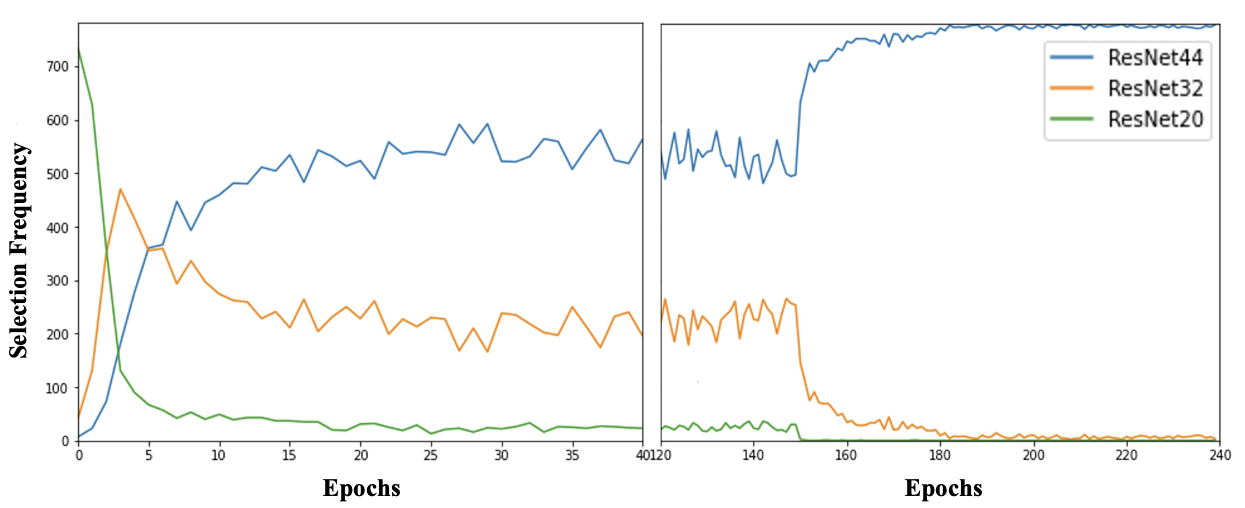}
    \end{tabular}
    }
% \vspace{-4mm}
   \caption{Frequency of being selected as a temporary teacher according to epochs.} 
   \label{fig:fig4}
\vspace{-2mm}   
\end{figure}

\textbf{Promotion Between Teacher and Student Group}
Before explaining the training process, we first talk about the requirements for promotion. To alleviate the problems of the previous study, we do not put all eggs in one basket as shown in Fig.~\ref{fig:overall} but divide the multiple networks into teacher group and student group. The teacher group includes the pivot teacher which is central to the overall training, and the student group includes immature teachers and students. The teacher group includes the pivot teacher who is central to overall training, and the student group includes immature teachers and students. Through Fig.~\ref{fig:fig3} (a), we judge that each network has different classes that show good performance, and through this, the qualification for teaching must be different depending on the data. The network that shows the best performance on the mini-batch image for each iteration is promoted from the student group to the teacher group with the qualification of a temporary teacher. When checking the total number of promotions per epoch as a temporary teacher, it can be seen that all networks are promoted frequently in Fig.~\ref{fig:fig4}. The reason is that each network has a different optimization speed, and a relatively shallow network can be optimized faster and more suitable for the role of a teacher.

%%%%%%%%%%%%%%%%%%%%%%%%%%%%%%%%%%%%%%%%%%%%%%%%%%%%%%%%%%%%%%%
%%%%%%%%%%%%%%%%%%%%%%%%%%%%%%%%%%%%%%%%%%%%%%%%%%%%%%%%%%%%%%%
%%%%%%%%%%%%%%%%%%%%%%%%%%%%%%%%%%%%%%%%%%%%%%%%%%%%%%%%%%%%%%%
%%%%%%%%%%%%%%%%%%%%%%%%%%%%%%%%%%%%%%%%%%%%%%%%%%%%%%%%%%%%%%% 
% 3가지 학습 단계
\subsection{Three Types of Teaching Methods in ORC}
From now on, we will describe the 3 training steps in ORC: Intensive teaching, Private teaching, Group teaching.

\textbf{Intensive Teaching.}
Because teacher should focus on the students not on itself, pivot teacher who holds the center needs to focus on samples that students find difficult and transfers corrected knowledge. The important point in this part is that the teacher needs to know what the students are having difficulty with. To find out where it is, we take feedback samples from student group. Students evaluate the performance of the samples for the mini-batch before training without updating parameters and then calculate the corresponding loss value, $L_{CE}$. The corresponding loss values are determined as the ratio of each student's feedback sample through softmax function. If the feedback sample consists of only mini-batch data, an overfitting problem may occur or a problem of deteriorating performance for classes that were originally predicted well may occur. To solve these problems, we use Mix-up to solve them. The data used for Mix-up is the mini-batch sample submitted as a feedback sample and the entire dataset, and the final feedback sample is created by fusing them. 
The feedback samples generated through the preceding process are used by the pivot teacher to catch intensively about the difficult part of the student group. Through intensive teaching, pivot teacher reduces the possibility of imparting false knowledge about samples that students struggle with. In addition, it can be seen that the problem of performance degradation for the well-predicted class previously mentioned is rather improved overall through Fig.~\ref{fig:fig3} (b). The loss of the intensive teaching $L_{I}$ for the pivot teacher's probability $P_{PT}$ is derived as follows:
\begin{equation}
\begin{split}
    %L_{PT} = \lambda L_{CE}(y_i\t, P(x_t)) + (1-\lambda)L_{CE}(y_\mathcal{F}, P(x_\mathcal{F})).
    L_{I} &= \mathcal{H}(\Tilde{y},P_{PT}(\Tilde{x})) \\
    &= \lambda \mathcal{H}(y_t, P_{PT}(\Tilde{x})) + (1-\lambda) \mathcal{H}(y_\mathcal{F}, P_{PT}(\Tilde{x})),
\end{split}
\label{eq:05}
\end{equation}
where $\Tilde{x}$,$\Tilde{y}$ are the feedback samples generated through the Mix-up.
and $\lambda$ is a randomly selected using the beta distribution.

\textbf{Private Teaching.}
In the next step, private teaching, the network with the highest performance in the student group is promoted as a temporary teacher in the teacher group. Students who are not promoted receive corrections for their tasks through group teaching, but student promoted as temporary teacher do not receive corrections. Therefore, the temporary teacher receives corrections, which is private teaching, from the pivot teacher to improve teaching ability. The loss function of the private teaching $L_P$ is represented as follows:
\begin{equation}
\begin{split}
    L_{P} = (1-\beta) L_{CE}(y_t,P_{TT}(x_t)) \\
    + \beta L_{KD}(P_{TT}(x_t),P_{PT}(x_t)),
\end{split}
\label{eq:06}
\end{equation}
where $P_{PT}(x_t)$ and $P_{TT}(x_t)$ are softened class probabilities of pivot teacher and temporary teacher, respectively.

\textbf{Group Teaching.}
Group-to-Group KD has been proposed in many forms~\cite{Quishan2020}\cite{DGKD}\cite{Mutual}\cite{one} but most methods have performed KD using all multiple networks at the same time. In this paper, for simplicity, all networks of the teacher group directly teach individual networks of the student group. Note that we have already constructed the teacher group with the correct distilled knowledge based on the previous 2 steps.
Group teaching loss $L^i_{G}$ for $i$th student network is defined as follows:
\begin{equation}
\begin{split}
     L^{i}_{G} = L_{KD}(P_{S_i}(x_t),P_{TT}(x_t)) \\
     + L_{KD}(P_{S_i}(x_t),P_{PT}(x_t)),
\end{split}
\label{eq:7}
\end{equation}
and the student total loss $L_{T}$ can be written as follows:
\begin{equation}
    L_{T} = \sum_{i} \{(1-\beta) L^{S_i}_{CE}(y_t,P_{S_i}(x_t)) + \beta L^{i}_{G}\}.
\label{eq:8}
\end{equation}

After group teaching, the temporary teacher is demoted from the teacher group to the student group. We repeat the proposed procedure to find a new temporary teacher in the next iteration. Through intensive teaching, the pivot teacher minimizes the possibility of transferring false knowledge about a sample that student group is difficult with, and through private teaching, temporary teacher receive corrections for false knowledge by improving their educational abilities. Finally, through group teaching, student group is corrected their incorrect knowledge.

%%%%%%%%%%%%%%%%%%%%%%%%%%%%%%%%%%%%%%%%%%%%%%%%%%%%%%%%%%%%%%%
%%%%%%%%%%%%%%%%%%%%%%%%%%%%%%%%%%%%%%%%%%%%%%%%%%%%%%%%%%%%%%%

%%%%%%%%%%%%%%%%%%%%%%%%%%%%%%%%%%%%%CIFAR10%%%%%%%%%%%%%%%%%%%%%%%%%%%%%%%%%%%
\begin{table}[t]
\begin{center}
\caption{Experimental results using ResNet on CIFAR-10 (Top-1 test accuracy);~\textbf{Bold} means the best accuracy and \underline{underline} is the second best. Teacher and student networks are ResNet26 (92.82\%) and ResNet8 (86.12\%), respectively. * means that we re-implement the method based on the paper. }
\vspace{-2mm}
\label{table:1}
\resizebox{\columnwidth}{!}{%
\begin{tabular}{ccccccc|c}
\hline
KD &FitNet &AT &FSP &BSS &DML &KDCL &\multirow{2}{*}{Ours} \\ 
\cite{HintonKD} &\cite{FitNets} &\cite{Zagoruyko2017}&\cite{FSP} &\cite{BSS} &\cite{Mutual} &\cite{Quishan2020}* &\\
\hline
\hline
86.66 &86.73 &86.86 &87.07 &87.32 &\underline{87.71} &87.48 &\textbf{88.76}  \\
\hline
\end{tabular}
}
\end{center}
\vspace{-6mm}
\end{table}
%%%%%%%%%%%%%%%%%%%%%%%%%%%%%%%%%%%%%%%%%%%%%%%%%%%%%%%%%%%%%%%%%%%%%%%%%%%%%

%%%%%%%%%%%%%%%%%%%%%%%%%%%%%%%%%%%%%%%%%%%%%%%%%%%%%%%%%%%%%%%%%
\begin{table*}[t]
\begin{center}
%\small
\caption{Comparison with KD methods based on the similar architectures; Top-1 accuracy (\%) on CIFAR-100. \textbf{Bold} is the best and \underline{underline} is the second best one.}
\label{table:2}
%\vspace{-3mm}
%\resizebox{\columnwidth}{!}{
\footnotesize
\begin{tabular}{c|c|c|c|c|c|c|c}
\hline
Teacher & WRN40-2 & WRN40-2 & ResNet56 & ResNet110 & ResNet110 & ResNet32$\times4 $&VGG13 \\
Student & WRN16-2 & WRN40-1 & ResNet20 & ResNet20 & ResNet32  & ResNet8$\times4$  &VGG8 \\
\hline
\hline
Teacher & 75.61 &75.61 &72.34 &74.31 &74.31 &79.42 &74.64\\
Student & 73.26  &71.98  &69.06 &69.06 &71.14 &72.50 &70.36\\
\hline
KD~\cite{HintonKD} &74.92  &73.54  &70.66 &70.67 &73.08 &73.33 &72.98\\
FitNet~\cite{FitNets} &73.58  &72.24  &69.21 &68.99 &71.06 &73.50 &71.02\\
AT~\cite{Zagoruyko2017}&74.08  &72.77  &70.55 &70.22 &72.31 &73.44  &71.43\\
SP~\cite{Tung2019}&73.83  &72.43  &69.67 &70.04 &72.69 &72.94  &72.68\\
CC~\cite{CC}&73.56  &72.21  &69.63 &69.48 &71.48 &72.97  &70.71\\
VID~\cite{VID}&74.11  &73.30  &70.38 &70.16 &72.61 &73.09   &71.23\\
RKD~\cite{RKD}&73.35  &72.22  &69.61 &69.25 &71.82 &71.90    &71.48\\
PKT~\cite{PKT}&74.54  &73.45  &70.34 &70.25 &72.61 &73.64   &72.88\\
AB~\cite{AB}&72.50  &72.38  &69.47 &69.53 &70.98 &73.17  &70.94\\
FT~\cite{FT}&73.25  &71.59  &69.84 &70.22 &72.37 &72.86 &70.58\\
FSP~\cite{FSP}&72.91  &0.00  &69.95 &70.11 &71.89 &72.62  &70.23\\
NST~\cite{NST}&73.68  &72.24  &69.60 &69.53 &71.96 & 73.30 &71.53\\
CRD~\cite{CRD}&75.48  &74.14 &71.16 &71.46&73.48& \underline{75.51} &73.94\\
SRRL~\cite{SRRL} &\underline{75.96}  &\underline{74.75}  &\underline{71.44} &\underline{71.51} &\underline{73.80} &\textbf{75.92} & \underline{74.40}\\
KDCL~\cite{Quishan2020} &67.73  &73.12  &70.58 &70.36 &72.67 & 74.03 &72.94\\
\hline
Ours& \textbf{76.4}&\textbf{75.34} &\textbf{72.07} &\textbf{71.6} &\textbf{74.46}& 75.00 &\textbf{74.68}\\
\hline
\end{tabular}
%}
\end{center}
\vspace{-6mm}
\end{table*}
%%%%%%%%%%%%%%%%%%%%%%%%%%%%%%%%%%%%%%%%%%%%%%%%%%%%%%%%%%%%%%%%%%%%%%%%%%%%%%%
%%%%%%%%%%%%%%%%%%%%%%%%%%%%%%%%%%%%%%%%%%%%%%%%%%%%
\begin{table*}[htp]
 \begin{center}
 \caption{Top-1 and Top-5 error rate (\%) on ImageNet~\cite{ImageNet}. Comparison results with the KD methods. We use the pivot teacher as ResNet34. The student group consists of ResNet28, ResNet22, and ResNet18. * denotes that we re-implemented the method based on the paper. }
 \label{table:3}
 %\vspace{-3mm}
 %\resizebox{\textwidth}{!}{
 \footnotesize
 \begin{tabular}{c|cc||c|c|c|c|c|c|c|c|c} 
 \hline
 &Teacher &Student &KD  &AT &SP &CC &ONE &CRD &DKD &KDCL &Ours \\
 &ResNet34 &ResNet18 &\cite{HintonKD} &\cite{Zagoruyko2017} &\cite{Tung2019} &\cite{CC} &\cite{CRD} &\cite{one} &\cite{dkd} &\cite{Quishan2020}*&   \\
 \hline
 \hline
 Top-1  &26.69 &30.25 &29.34 &29.30 &29.38 &30.04 &29.45 &28.83  &\underline{28.30}  &29.57 & \textbf{28.00}\\       
 Top-5  &8.58 &10.93 &10.12 &10.00 &10.20 &10.83 &10.41 &9.87 &\underline{9.59}  &10.01 &  \textbf{9.13}\\
 \hline
 \end{tabular}
 %}
\end{center}
\vspace{-4mm}
\end{table*}
%%%%%%%%%%%%%%%%%%%%%%%%%%%%%%%%%%%%%%%%%%%%%%%%%%%%%%%%%%%%%%%%%%%%%%%%

\section{Experiment Settings}
In this paper, we evaluate and compare our proposed method to well-known KD approaches using the classification benchmark datasets such as CIFAR~\cite{CIFAR} and ImageNet~\cite{ImageNet} following the well-known protocols~\cite{CRD}. We implement ours by PyTorch~\cite{Paszke2017} using eight V100 GPUs. 

\textbf{Datasets.} CIFAR~\cite{CIFAR} is divided into 10 classes and 100 classes.(i.e., CIFAR-10 and CIFAR-100) There are 60,000 images of 32$\times$32 pixel resolution, consisting of 50,000 images for training and 10,000 images for testing. ImageNet~\cite{ImageNet}, a popularly used dataset for the image classification, contains 1.28M images of 224$\times$224 pixel resolution for training and 50,000 images for validation, with 1,000 classes.

\textbf{Networks.} In order to increase the fairness of the experiment, we employ networks based on the knowledge distillation experiment of CRD~\cite{CRD}, which has been used in many studies. The networks are used as follows: ResNet~\cite{ResNet,ResNet1}, WRN~\cite{WideResNet}, ShufflenetV1~\cite{zhang2018shufflenet}, ShufflenetV2~\cite{ma2018shufflenet}, Mobilenet~\cite{howard2017mobilenets}, and VGG~\cite{vgg}.

\textbf{Implementation details.} (1) \textbf{CIFAR:} We use common data augmentations such as random crop and horizontal flip. We use a stochastic gradient descent optimizer with momentum 0.9 and weight decay 0.0005. We generally initialize the learning rate to 0.05 but set it to 0.1 for the MobilenetV2, ShufflenetV1, and ShufflenetV2 networks. We then decrease it by 0.1 at epochs 100, 150, 210 until the last 240 epochs. We set a batch size to 64 and a temperature $\tau$ to 4. 
(2) \textbf{ImageNet:} For ImageNet, we employ a stochastic gradient descent optimizer with nesterov momentum 0.9, weight decay 2e-5. We set the batch size to 64, the temperature $\tau$ to 2, the base learning rate to 0.1, and decrease the learning rate by 0.1 every 30 epochs for a total of 3 times.

\subsection{Comparison with State-Of-The-Art methods}
On the CIFAR-10 and CIFAR-100~\cite{CIFAR} and ImageNet~\cite{ImageNet} dataset, we compare our method against previous knowledge distillation methods proposed from the past to the present.

% CIFAR-10
\textbf{Benchmark on CIFAR-10.} We use four ResNets: e.g., ResNet26, ResNet20, ResNet14, and ResNet8 for multiple teacher-based KD methods including both ours and KDCL. 
%ResNet26 is the teacher and ResNet8 is the target student network. 
As shown in Table~\ref{table:1}, we compare ours with the Hinton's KD~\cite{HintonKD}, FitNet~\cite{FitNets}, AT~\cite{Zagoruyko2017}, FSP~\cite{FSP}, BSS~\cite{BSS}, DML~\cite{Mutual}, and KDCL~\cite{Quishan2020} methods. We observe that the multiple teacher-based KD, e.g., KDCL, does not always achieve the best accuracy compared with the previous methods, e.g., DML. However, ours results in the best accuracy compared with the other KD methods, which validates that our ORC effectively uses the multiple teachers for KD. 

% CIFAR-100
\textbf{Benchmark on CIFAR-100.} We follow the CRD~\cite{CRD} experimental protocol to verify the generality of the proposed method compared with the fifteen previous works. We also use the multiple networks\footnote{The detail could be found in Supplementary material} for ours and KDCL.
Table~\ref{table:2} shows an experiment where the seven similar backbone architectures between the student and the teacher are used for KD. We confirm that our method achieved the best accuracy in all results based on the similar architectures from WRN~\cite{WideResNet} to ResNet~\cite{ResNet} except the KD result from ResNet32$\times$4 to ResNet8$\times$4. Note that, as shown by the performances of KDCL, we observed that using multiple networks does not always result in better performances, and rather how we use multiple networks is important. Furthermore, Table~\ref{table:4} summarizes the comparison results of KD between the different network architecture types. Our method achieves the best accuracy in four out of six experiments and the second best accuracy in two experiments. From this result, we can conclude that our method performs good performance no matter which architecture is used for KD. 

% ImageNet
\textbf{Benchmark on ImageNet.} 
For proving the scalability of the proposed method to the large-scale dataset, i.e., ImageNet~\cite{ImageNet}, we validate the superiority of our method compared with the well-known KD methods in Table~\ref{table:3}. We basically re-implement the multiple network-based online KD methods using ResNet34, ResNet28, ResNet22, and ResNet18 for ours and KDCL~\cite{Quishan2020}. Our method achieves better performance than the multiple network-based KD (e.g., KDCL~\cite{Quishan2020}) and others from Top-1 to Top-5 error rates in ImageNet validation.

%%%%%%%%%%%%%%%%%%%%%%%%%%%%%%%%%% Heterogenous %%%%%%%%%%%%%%%%%%%%%%%%%%%%%%%
\begin{table*}[htp]
\begin{center}
\caption{Comparison with KD methods based on the different architectures; Top-1 accuracy (\%) on CIFAR-100. \textbf{Bold} is the best and \underline{underline} is the second best one. 
%\textbf{Bold} is the best and \underline{underline} is the second best one. We denote by * method where we re-implement the method based on the paper.
}
%\vspace{-3mm}
\label{table:4}
%\resizebox{\textwidth}{!}{
\footnotesize
\begin{tabular}{c|c|c|c|c|c|c}
\hline
Teacher &VGG13 &ResNet50   &ResNet50 &ResNet32$\times4 $ &ResNet32$\times4 $    &WRN40-2 \\
Student &MobilenetV2 &MobilenetV2 &VGG8 &ShufflenetV1   &ShufflenetV2  &ShufflenetV1 \\
\hline
\hline
Teacher & 74.64 &79.34 &79.34 &79.42 &79.42 &75.61\\
Student & 64.60  &64.60  &70.36 &70.50 &71.82 &70.50\\
\hline
KD~\cite{HintonKD} &67.37  &67.35  &73.81 &74.07 &74.45 &74.83\\
FitNet~\cite{FitNets} &64.14  &63.16  &70.69 &73.59 &73.54 &73.73\\
AT~\cite{Zagoruyko2017}&59.40  &58.58  &71.84 &71.73 &72.73 &73.32\\
SP~\cite{Tung2019}&66.30  &68.08 &73.34 &73.48 &74.56 &74.52\\
CC~\cite{CC}&64.86  &65.43  &70.25 &71.14 &71.29 &71.38\\
VID~\cite{VID}&65.56  &67.57  &70.30 &73.38 &73.40 &73.61\\
RKD~\cite{RKD}&64.52  &64.43  &71.50 &72.28 &73.21 &72.21\\
PKT~\cite{PKT}&67.13  &66.52  &73.01 &74.10 &74.69 &73.89\\
AB~\cite{AB}&66.06  &67.20  &70.65 &73.55 &74.31 &73.34\\
FT~\cite{FT}&61.78  &60.99  &70.29 &71.75 &72.50 &72.03\\
NST~\cite{NST}&58.16  &64.96  &71.28 &74.12 &74.68 &74.89\\
CRD~\cite{CRD}&\textbf{69.73}  &69.11  &74.30 &75.11 &75.65 &76.05\\
SRRL~\cite{SRRL} &69.14  &\underline{69.45}  &\underline{74.46} &\underline{75.66} &\underline{76.40} & \textbf{76.61}\\
KDCL~\cite{Quishan2020} &67.45  &67.64  &73.03 &74.32 &75.35 & 74.79\\
\hline
Ours& \underline{69.70}&\textbf{69.63} &\textbf{74.65} &\textbf{76.60} &\textbf{76.84}&\underline{76.56}\\
\hline
\end{tabular}
%}
\end{center}
\vspace{-4mm}
\end{table*}
%%%%%%%%%%%%%%%%%%%%%%%%%%%%%%%%%%%%%%%%%%%%%%%%%%%%%%%%%%%%%%%%%%%%%%%%%%%%%%%

\subsection{Ablation Studies}
In this section, we have done all experiments for the ablation test with ResNet56 (i.e., pivot teacher), ResNet44, ResNet32, and ResNet20 (i.e., target) in the CIFAR-100. The baselines of this section are the results of learning the individual networks by themselves.

\textbf{Individual components of the proposed method.} Table~\ref{table:5} summarizes the performance gain according to the different components of the proposed method. The proposed ORC-based group teaching shows better results compared with the baseline networks. When we add the intensive teaching with the group teaching, the performances of the multiple networks including the pivot teacher are improved individually. It means that intensive teaching with the error images for feedback from the student group helps the pivot teacher prevent transferring false knowledge of samples which students struggle with. Finally, when private teaching is added, performance improvement can be also confirmed, which shows that the false knowledge of the temporary teacher is corrected through private teaching and the delivery of defective information to the student group is prevented.

%%%%%%%%%%%%%%%%%%%%%%%%%%%%%%%%%%%%%%%%%%%%%%%%%%%%%%%
\begin{table}[t]
\begin{center}
\caption{Ablation results by the individual components of the proposed method on CIFAR-100.}
\label{table:5}
% \vspace{-2mm}
\resizebox{\columnwidth}{!}{
\begin{tabular}{ccc|c|c|c|c}
\\
\hline
Group & Intensive & Private& ResNet56 & ResNet44 & ResNet32 & ResNet20 \\ 
Teaching& Teaching& Teaching& (pivot teacher) & (network1) & (network2) & (network3)\\
\hline
\hline
\multicolumn{3}{c|}{Baseline} &72.34&70.53 &70.11  &69.53\\
\hline

 \checkmark &            &             & -             &72.68          &73.46          &71.33\\
 \checkmark & \checkmark &             &73.68          &73.36          &73.79          &71.59\\
 \checkmark & \checkmark & \checkmark  &\textbf{74.55} &\textbf{75.17} &\textbf{74.25} &\textbf{72.07}\\
 
\hline
          
\end{tabular}   
}
\vspace{-6mm}
\end{center}
\end{table}
%%%%%%%%%%%%%%%%%%%%%%%%%%%%%%%%%%%%%%%%%%%%%%%%%%%%%%%%%%%%%%%%%%%%%%

% 앙상블 vs 개별
\textbf{Style of group teaching.} Here, we investigate how the teacher group should teach the student group effectively. The ensemble method is directly inspired by KDCL~\cite{Quishan2020} and teaches the student group after averaging the logits of networks in the teacher group. On the other hand, in the individual method, the teacher group's network separately teaches the student group's network. Table~\ref{table:6} summarizes the comparison results of the ensemble and the individual method, and the individual method clearly outperforms the former. Because in the case of KDCL, all networks are used for teaching and averaging the logits is necessary to prevent incorrect knowledge from the teacher networks. However, in our method, because of the proposed ORC, only networks with good knowledge always belong to the teacher group, so when we teach the student group individually, it is unlikely that false knowledge might be distilled from the teacher group as shown in Fig.~\ref{fig:fig1} (b).

\textbf{Online role change works well.} We confirmed through Fig.~\ref{fig:fig4} that deep networks do not always show good performance because the optimization speed is different for each depth of the model and the classes that each network predicts well are different. To support the content, we fixed the deep model as a temporary teacher and conducted a comparative experiment with ORC. Through Table~\ref{table:7}, it can be confirmed that continuously using a network suitable for the teacher's role is superior in performance. Additionally, the results of comparison with DGKD, which conducts knowledge transfer after pre-training all teachers, can also be found in Table~\ref{table:7}, and it can be confirmed that ORC performs better despite being an online manner training. Through the previous two experiments, we prove that it is an efficient algorithm that obtains better performance by reducing false knowledge from the student group through efficient online role change of networks.

%%%%%%%%%%%%%%%%%%%%%%%%%%%%%%%%%%%%%%%%%%%%%%%%%%%%%%%%%
\begin{table}[t]
\begin{center}
\caption{Comparison of accuracy according to style in group teaching; ensemble logit means teaching by averaging the logits of the pivot teacher and temporary teacher, and individual logit means teaching each logit independently.}
\label{table:6}
% \vspace{-2mm}
\resizebox{\columnwidth}{!}{
\begin{tabular}{c|c|c|c|c}
\hline
Group teaching& ResNet56	&ResNet44 &ResNet32 &ResNet20\\
style & (pivot teacher) & (network1) & (network2) & (network3)\\
\hline
\hline
Baseline & 72.34 &70.53 & 70.11 & 69.53\\
\hline
Ensemble logit & 73.72 &74.55 &73.78 &71.56\\
Individual logit & \textbf{74.55} &\textbf{75.17} &\textbf{74.25} &\textbf{72.07}\\
\hline
\end{tabular}
}
\end{center}
\vspace{-6mm}
\end{table}
%%%%%%%%%%%%%%%%%%%%%%%%%%%%%%%%%%%%%%%%%%%%%%%%%%%%%%%%%%%%%%%%%%%%%%%%%%%%%%%%%%

%
\textbf{Online KD using multiple networks.} We compared ORC's performance improvement with multiple newtwork-based online KD methods in Table~\ref{table:8} such as DML~\cite{Mutual} and KDCL~\cite{Quishan2020} to show that the performance improvement is not simply due to the multiple-teacher-based method. Note that multiple networks all have the same number of layers in the same architecture. All methods achieved better than the baseline, but the performance improvement rate is different depending on the network architecture used. However, our method always performs best accuracy from ResNet32$\times$4 to WRN40-2. 

\textbf{Performance comparison of the Mix-Up methods.} The performance comparison of the different inputs for MixUp used in intensive teaching is summarized in Table~\ref{table:9}. When we use the feedback samples, $x_\mathcal{F}$, without MixUp for the intensive teaching, we can observe some improvements compared with the baseline and it means that the feedback composed of error examples from the students provides clues as to how the pivot teacher teaches the student group more effectively. 
However, due to potential overlap among the feedback samples from students, the number of distinct types of feedback samples is less than the total number of training examples. Therefore, we adopted the MixUp with both feedback samples and training instances to augment the data. Note that when we use the MixUp only for feedback examples, the overfitting problem could occur because the teacher network is biased to only error samples, not the whole data. Through our mix-up method, we confirm that the performance is improved overall as shown in Figure~\ref{fig:fig3} (b) rather than the overfitting problem of the pivot teacher.

%%%%%%%%%%%%%%%%%%%%%%%%%%%%%%%%%%%%%%%%%%%%%%%%%%%%%%%%%%%%%
\begin{table}[t]
\begin{center}
\caption{Comparison of the performance of the teacher group consisting of fixed teachers and the teacher group with temporary teachers. * denotes that the models are pretrained and not fine-tuned.}
\label{table:7}
% \vspace{-2mm}
\resizebox{\columnwidth}{!}{%
\begin{tabular}{c|ccc}
\hline
\# of Pivot teacher &P=1(ORC) &P=2(w/o ORC) &P=3(DGKD)\\
\hline
\hline
ResNet56 &\textbf{74.55} &73.78 &72.34*\\
ResNet44 &\textbf{75.17} &73.09 &70.53*\\
ResNet32 &\textbf{74.25} &73.63 &70.11*\\
ResNet20 &\textbf{72.07} &71.36 &71.92\\
\hline
\end{tabular}   
}
\end{center}
\vspace{-4mm}
\end{table}
%%%%%%%%%%%%%%%%%%%%%%%%%%%%%%%%%%%%%%%%%%%%%%%%%%%%%%%%%%%%%%%%%%%%%%%%%%%%%%%%%%
%%%%%%%%%%%%%%%%%%%%%%%%%%%%%%%%%%%%%%%%%%%%%%%%%%%%%%%%%%%%%
\begin{table}[t]
\begin{center}
\caption{Comparison with online KD methods using multiple networks on CIFAR-100; The number in the parentheses is the number of the networks used for KD. We all use the same size networks.}
% \vspace{-2mm}
\label{table:8}
\resizebox{\columnwidth}{!}{%
\begin{tabular}{c|c|c|c|c|c|c} 
\hline
 Method(\#) & ResNet32$\times4$ &ResNet110 &ResNet56 &ResNet50 &VGG13 &WRN40-2 \\
\hline
\hline
Baseline &79.42 &74.31 &72.34 &79.34 &74.31 &75.61\\
\hline
KD(2) &80.25 &\underline{76.60} &74.87 &80.01 &76.15 &78.07\\
DML(2) & \underline{80.57} & 74.47 & 74.47& 80.74 &\underline{77.34} & 77.63\\
DML(4) &80.15 &76.28 &\underline{75.17} &80.14 &76.96 &\underline{78.22}\\
KDCL(4)&80.23 &76.23 &74.79 &\underline{80.87} &76.89 &78.03\\
\hline
Ours(4) &\textbf{81.56} &\textbf{78.03} &\textbf{76.54} &\textbf{81.21} &\textbf{77.45}& \textbf{79.39} \\
\hline
\end{tabular}
}
\end{center}
\vspace{-6mm}
\end{table}
%%%%%%%%%%%%%%%%%%%%%%%%%%%%%%%%%%%%%%%%%%%%%%%%%%%%%%%%%%%%%%%%%%%%%%%%%%%%%%%%%%

%
\textbf{The number of temporary teachers.} We show how many temporary teachers are effective for KD through an experiment in Table~\ref{table:10}. We set the temporary teacher from 0 to $k$ and prove it by comparing the performance of all networks. It shows the best performance when set to $k$ = 1. Compared to the baseline, ResNet44, ResNet32, and ResNet20 gain 4.64\%, 4.14\%, and 2.54\% improvements, respectively.
When there is no temporary teacher (e.g., $k$=0), the performance improvement is not significant. It means that there is a limit to performance improvement with only the pivot teacher. When $k$=2, the performance is better than when $k$=0, but not as good as when $k$=1. Because many networks (e.g., when $k=2$, 66\% networks are promoted from the student group) in the student group are promoted to the temporary teachers, the false knowledge can be transferred from the teacher group to the student group, as shown in Fig.~\ref{fig:fig1} (a).

\section{Conclusion}
In this paper, we propose a novel mechanism called Network group-based KD using ORC strategy. The previous KD methods based on multiple teachers had limitations in that immature teachers distill and transfer false knowledge. In order to overcome this issue, we promote a top-ranked student to a temporary teacher in the student group at every iteration. Then, group teaching is performed after including the temporary teacher in the teacher group to prevent false knowledge transfer. Furthermore, we propose intensive teaching in which the student group provides flexible feedback on error instances to fine-tune the pivot teacher. This mechanism enables the pivot teacher to understand the student group deeper. In addition, through private teaching, temporary teacher can perform the role of education well. Our method is used at each iteration of the training process, and it has been demonstrated by experiments to be an effective way of knowledge transfer.

%%%%%%%%%%%%%%%%%%%%%%%%%%% Mix-up에 대한 분석 %%%%%%%%%%%%%%%%%%%%%%%%%%%%
\begin{table}[t]
\begin{center}
\caption{Performance comparison of the MixUp methods using the different examples for the intensive teaching on CIFAR-100.}
\label{table:9}
% \vspace{-2mm}
\resizebox{\columnwidth}{!}{%
\begin{tabular}{c|c|c|c|c}
\hline
MixUp inputs& ResNet56 & ResNet44 & ResNet32 & ResNet20 \\ 
 & (pivot teacher) & (network1) & (network2) & (network3)\\
\hline
\hline
{Baseline} &72.34&70.53 &70.11  &69.53\\
\hline
%No MixUp  
No MixUp ($x_\mathcal{F}$) & 73.63 &74.52 &73.87 & 71.38\\
%$x_i$ only  &74.07 &74.96 &74.13 & 71.03\\ 
$x_\mathcal{F}$ only &{73.59} & {74.79}&\textbf{74.55}&{71.43}\\
% $x_\mathcal{F}$ only & \textcolor{blue}{73.59} & \textcolor{red}{75.34}& \textcolor{red}{73.31}&\textcolor{red}{71.43}\\
Ours ($x_\mathcal{F}$ and $x_t$) &\textbf{74.55} &\textbf{75.17} &{74.25} &\textbf{72.07}\\
\hline
\end{tabular}   
}
\vspace{-4mm}
\end{center}
\end{table}
%%%%%%%%%%%%%%%%%%%%%%%%%%%%%%%%%%%%%%%%%%%%%%%%%%%%%%%%%%%%%%%%%%%%%%

%%%%%%%%%%%%%%%%%%%%%%%%%%%%%%%%%%%%%%%%%%%%%%%%
\begin{table}[t]
\begin{center}
\caption{Comparing the performance of all networks in ResNet with $k$ temporary teachers; $k$ is the number of temporary teachers used for KD.}
\resizebox{\columnwidth}{!}{
\label{table:10}
% \vspace{-2mm}
\footnotesize
%\resizebox{\columnwidth}{!}{%
\begin{tabular}{c|c|c|c|c}
\\
\hline
\multirow{2}{*}{$k$} &ResNet56	&ResNet44 &ResNet32 &ResNet20\\
&(pivot teacher) &(network1) &(network2) &(network3)\\
\hline
\hline
Baseline & 72.34 &70.53 & 70.11 & 69.53\\
\hline
0 & 73.78 &75.03 &73.81 &70.98\\
1 & \textbf{74.55} &\textbf{75.17} &\textbf{74.25} & \textbf{72.07}\\
2 & 74.07 &75.12 &74.06 &71.22\\
\hline
\end{tabular}
}
\end{center}
\vspace{-6mm}
\end{table}
%%%%%%%%%%%%%%%%%%%%%%%%%%%%%%%%%%%%%%%%%%%%%%%%%%%%%%%%%%%%%%%%%%%%%%

% 

\noindent\textbf{Acknowledgement.} This work was supported by Korea IITP grants (No.2021-0-00951, Dev. of Cloud based Autonomous Driving AI learning SW; No.2021-0-02068, AI Innovation Hub; IITP-2023-No.RS-2023-00255968, AI Convergence Innovation Human Resources Dev.) and by Korea NRF grant (NRF-2022R1A2C1091402). W. Hwang is the corresponding author.

{\small
% \bibliographystyle{ieee_shortname}%ieee_fullname}
% \bibliography{egbib}

\begin{thebibliography}{10}\itemsep=-1pt

\bibitem{VID}
S.~Ahn, S.~X., Hu, A.~Damianous, N.~D. Lawrence, and Z.~Dai.
\newblock Variational information distillation for knowledge transfer.
\newblock {\em IEEE Conf. on Computer Vision and Pattern Recognition}, Jun.
  2019.

\bibitem{Defang2020}
D.~Chen, J.~Mei, C.~Wang, Y.~Feng, and C.~Chen.
\newblock Online knowledge distillation with diverse peers.
\newblock {\em 34th AAAI Conf. on Artificial Intelligence}, Feb. 2020.

\bibitem{chen2016quantized}
W.~Chen, J.~Wilson, S.~Tyree, K.~Q. Weinberger, and Y.~Chen.
\newblock Compressing convolutional neural networks in the frequency domain.
\newblock In {\em Proceedings of the 22nd ACM SIGKDD International Conference
  on Knowledge Discovery and Data Mining}, pages 1475--1484, 2016.

\bibitem{itkd}
H.~Cho, J.~Choi, G.~Baek, and W.~Hwang.
\newblock itkd: Interchange transfer-based knowledge distillation for 3d object
  detection.
\newblock In {\em Proceedings of the IEEE/CVF Conference on Computer Vision and
  Pattern Recognition}, pages 13540--13549, 2023.

\bibitem{cho2019}
J.~Cho and B.~Hariharan.
\newblock On the efficacy of knowledge distillation.
\newblock {\em IEEE International Conf. on Computer Vision}, pages 4794--4802,
  Oct. 2019.

\bibitem{cho2021deep}
J.~Cho, D.~Min, Y.~Kim, and K.~Sohn.
\newblock Deep monocular depth estimation leveraging a large-scale outdoor
  stereo dataset.
\newblock {\em Expert Systems with Applications}, 178:114877, 2021.

\bibitem{ImageNet}
J.~Deng, W.~Dong, R.~Socher, L.-J. Li, K.~Li, and L.~Fei-Fei.
\newblock Imagenet: A large-scale hierarchical image database.
\newblock {\em IEEE Conf. on Computer Vision and Pattern Recognition}, Jun.
  2009.

\bibitem{Quishan2020}
Q.~Guo, X.~Wang, Y.~Wu, Z.~Yu, D.~Liang, X.~Hu, and P.~Luo.
\newblock Online knowledge distillation via collaborative learning.
\newblock {\em IEEE Conf. on Computer Vision and Pattern Recognition}, pages
  4320--4328, Jun. 2020.

\bibitem{han2015deeppruning}
S.~Han, H.~Mao, and W.~Dally.
\newblock Deep compression: Compressing deep neural networks with pruning,
  trained quantization and huffman coding.
\newblock {\em arXiv preprint arXiv:1510.00149}, 2015.

\bibitem{han2015learningpruning}
S.~Han, J.~Pool, J.~Tran, and W.~Dally.
\newblock Learning both weights and connections for efficient neural network.
\newblock {\em Advances in neural information processing systems}, 28, 2015.

\bibitem{ResNet}
K.~He, X.~Zhang, S.~Ren, and J.~Sun.
\newblock Deep residual learning for image recognition.
\newblock {\em IEEE Conf. on Computer Vision and Pattern Recognition}, pages
  770--778, Jun. 2016.

\bibitem{ResNet1}
K.~He, X.~Zhang, S.~Ren, and J.~Sun.
\newblock Identity mappings in deep residual networks.
\newblock In {\em European conference on computer vision}, pages 630--645.
  Springer, 2016.

\bibitem{he2018pruning}
Y.~He, J.~Lin, Z.~Liu, H.~Wang, L.~Li, and S.~Han.
\newblock Amc: Automl for model compression and acceleration on mobile devices.
\newblock In {\em Proceedings of the European conference on computer vision
  (ECCV)}, pages 784--800, 2018.

\bibitem{BSS}
B.~Heo, M.~Lee, S.~Yun, and J.~Choi.
\newblock Improving knowledge distillation with supporting adversarial samples.
\newblock {\em 33rd AAAI Conf. on Artificial Intelligence}, pages 3771--3778,
  Feb. 2019.

\bibitem{AB}
B.~Heo, M.~Lee, S.~Yun, and J.~Choi.
\newblock Knowledge transfer via distillation of activation boundaries formed
  by hidden neurons.
\newblock {\em 34th AAAI Conf. on Artificial Intelligence}, pages 3779--3787,
  Jan. 2019.

\bibitem{HintonKD}
G.~Hinton, O.~Yinyals, and J.~Dean.
\newblock Distillation the knowledge in a neural network.
\newblock {\em arXiv preprint arXiv:1503.02531}, Mar. 2015.

\bibitem{howard2017mobilenets}
A.~Howard, M.~Zhu, B.~Chen, D.~Kalenichenko, W.~Wang, T.~Weyand, M.~Andreetto,
  and H.~Adam.
\newblock Mobilenets: Efficient convolutional neural networks for mobile vision
  applications.
\newblock {\em arXiv preprint arXiv:1704.04861}, 2017.

\bibitem{NST}
Z.~Huang and N.~Wang.
\newblock Like what you like: Knowledge distill via neuron selectivity
  transfer.
\newblock {\em arXiv preprint arXiv:1707.01219}, 2017.

\bibitem{hubara2016quantized}
I.~Hubara, M.~Courbariaux, D.~Soudry, R.~El-Yaniv, and Y.~Bengio.
\newblock Binarized neural networks.
\newblock {\em Advances in neural information processing systems}, 29, 2016.

\bibitem{FT}
J.~Kim, S.~Park, and N.~Kwak.
\newblock Paraphrasing complex network: Network compression via factor
  transfer.
\newblock {\em Advances in Neural Information Processing Systems}, pages
  2760--2769, Dec. 2018.

\bibitem{CIFAR}
A.~Krizhevsky.
\newblock Learning multiple layers of features from tiny images.
\newblock {\em Tech. Rep}, Apr. 2009.

\bibitem{one}
X.~Lan, X.~Zhu, and S.~Gong.
\newblock Knowledge distillation by on-the-fly native ensemble.
\newblock {\em Advances in Neural Information Processing Systems}, pages
  7528--7538, Dec. 2018.

\bibitem{li2016pruning}
H.~Li, A.~Kadav, I.~Durdanovic, H.~Samet, and H.~Graf.
\newblock Pruning filters for efficient convnets.
\newblock {\em arXiv preprint arXiv:1608.08710}, 2016.

\bibitem{ma2018shufflenet}
N.~Ma, X.~Zhang, H.~Zheng, and J.~Sun.
\newblock Shufflenet v2: Practical guidelines for efficient cnn architecture
  design.
\newblock In {\em Proceedings of the European conference on computer vision
  (ECCV)}, pages 116--131, 2018.

\bibitem{TAKD2020}
S.~I. Mirzadeh, M.~Farajtabar, A.~Li, N.~Levine, A.~Matsukawa, and
  H.~Ghasemzadeh.
\newblock Improved knowledge distillation via teacher assistant.
\newblock {\em 34th AAAI Conf. on Artificial Intelligence}, pages 5191--5198,
  Feb. 2020.

\bibitem{RKD}
W.~Park, D.~Kim, Y.~Lu, and M.~Cho.
\newblock Relational knowledge distillation.
\newblock {\em IEEE Conf. on Computer Vision and Pattern Recognition}, Jun.
  2019.

\bibitem{Paszke2017}
A.~Paszke, S.~Gross, S.~Chintala, and et. al.
\newblock Automatic differentiation in pytorch.
\newblock {\em NIPS Autodiff Workshop}, 2017.

\bibitem{PKT}
D.~Pathak, P.~Kahenbuhl, J.~Donahue, T.~Darrell, and A.~Efros.
\newblock Context encoders: Feature learning by inpainting.
\newblock {\em IEEE Conf. on Computer Vision and Pattern Recognition}, Jun.
  2016.

\bibitem{CC}
B.~Peng, X.~Jin, J.~Liu, D.~Li, Y.~Wu, Y.~Liu, S.~Zhou, and Z.~Zhang.
\newblock Correlation congruence for knowledge distillation.
\newblock {\em IEEE International Conf. on Computer Vision}, Oct. 2019.

\bibitem{rastegari2016quantizedr}
M.~Rastegari, V.~Ordonez, J.~Redmon, and A.~Farhadi.
\newblock Xnor-net: Imagenet classification using binary convolutional neural
  networks.
\newblock In {\em European conference on computer vision}, pages 525--542.
  Springer, 2016.

\bibitem{FitNets}
A.~Romero, N.~Ballas, S.~Kahou, A.~Chassang, C.~Gatta, and Y.~Bengio.
\newblock Fitnets: Hints for thin deep nets.
\newblock {\em International Conf. on Learning Representations}, 2015.

\bibitem{vgg}
K.~Simonyan and A.~Zisserman.
\newblock Very deep convolutional networks for large-scale image recognition.
\newblock {\em arXiv preprint arXiv:1409.1556}, 2014.

\bibitem{DGKD}
W.~Son, J.~Na, J.~Choi, and W.~Hwang.
\newblock Densely guided knowledge distillation using multiple teacher
  assistants.
\newblock In {\em Proceedings of the IEEE/CVF International Conference on
  Computer Vision}, pages 9395--9404, 2021.

\bibitem{CRD}
Y.~Tian, D.~Krishnan, and P.~Isola.
\newblock Contrastive representation distillation.
\newblock {\em International Conf. on Learning Representations}, Apr. 2020.

\bibitem{Tung2019}
F.~Tung and G.~Mori.
\newblock Similarity-preserving knowledge distillation.
\newblock {\em IEEE International Conf. on Computer Vision}, pages 1365--1374,
  Oct. 2019.

\bibitem{wu2016quantized}
J.~Wu, C.~Leng, Y.~Wang, Q.~Hu, and J.~Cheng.
\newblock Quantized convolutional neural networks for mobile devices.
\newblock In {\em Proceedings of the IEEE conference on computer vision and
  pattern recognition}, pages 4820--4828, 2016.

\bibitem{SELF2020}
G.~Xu, Z.~Liu, X.~Li, and C.~C. Loy.
\newblock Knowledge distillation meets self-supervision.
\newblock {\em European Conf. on Computer Vision}, Aug. 2020.

\bibitem{SRRL}
J.~Yang, B.~Martinez, A.~Bulat, and G.~Tzimiropoulos.
\newblock Knowledge distillation via softmax regression representation
  learning.
\newblock In {\em International Conference on Learning Representations}, 2021.

\bibitem{FSP}
J.~Yim, D.~Joo, J.~Bae, and J.~Kim.
\newblock A gift from knowledge distillation: Fast optimization, network
  minimization and transfer learning.
\newblock {\em IEEE Conf. on Computer Vision and Pattern Recognition}, Jul.
  2017.

\bibitem{WideResNet}
S.~Zagoruyko and N.~Komodakis.
\newblock Wide residual networks.
\newblock {\em British Machine Vision Conference}, pages 87.1--87.12, Sept.
  2016.

\bibitem{Zagoruyko2017}
S.~Zagoruyko and N.~Komodakis.
\newblock Paying more attention to attention: Improving the performance of
  convolutional neural networks via attention transfer.
\newblock {\em International Conf. on Learning Representations}, May 2017.

\bibitem{zhang2018shufflenet}
X.~Zhang, X.~Zhou, M.~Lin, and J.~Sun.
\newblock Shufflenet: An extremely efficient convolutional neural network for
  mobile devices.
\newblock In {\em Proceedings of the IEEE conference on computer vision and
  pattern recognition}, pages 6848--6856, 2018.

\bibitem{Mutual}
Y.~Zhang, T.~Xiang, T.~Hospedales, and H.~Lu.
\newblock Deep mutual learning.
\newblock {\em IEEE Conf. on Computer Vision and Pattern Recognition}, pages
  4320--4328, Jun. 2018.

\bibitem{dkd}
B.~Zhao, Q.~Cui, R.~Song, Y.~Qiu, and J.~Liang.
\newblock Decoupled knowledge distillation.
\newblock In {\em Proceedings of the IEEE/CVF Conference on computer vision and
  pattern recognition}, pages 11953--11962, 2022.

\end{thebibliography}

}

\end{document}